\documentclass[10pt,twocolumn,letterpaper]{article}

\usepackage{cvpr}
\usepackage{times}
\usepackage{epsfig}
\usepackage{graphicx}
\usepackage{amsmath}
\usepackage{amssymb}
\usepackage{multirow}

\usepackage[breaklinks=true,bookmarks=false]{hyperref}

\newif\ifsubmission\submissionfalse
\ifsubmission
\else
\cvprfinalcopy % *** Uncomment this line for the final submission
\fi

\def\cvprPaperID{7425} % *** Enter the CVPR Paper ID here

\begin{document}

%%%%%%%%% TITLE
\title{Creating High Resolution Images with a Latent Adversarial Generator}

\author{%
  David Berthelot \quad Peyman Milanfar\\
  Google Research \\
  \texttt{\{dberth,milanfar\}@google.com} \\
  \and
  Ian Goodfellow \\
  Work done at Google \\
  \texttt{ian-academic@mailfence.com}}

\maketitle
%\thispagestyle{empty}

%%%%%%%%% ABSTRACT
\begin{abstract}
   Generating realistic images is difficult, and many formulations for this task have been proposed recently. If we restrict the task to that of generating a particular class of images, however, the task becomes more tractable. That is to say, instead of generating an arbitrary image as a sample from the manifold of natural images, we propose to sample images from a particular "subspace" of natural images, directed by a low-resolution image from the same subspace. 
   
   The problem we address, while close to the formulation of the single-image super-resolution problem, is in fact rather different. Single image super-resolution is the task of predicting the image closest to the ground truth from a relatively low resolution image. We propose to produce samples of high resolution images given extremely small inputs with a new method called Latent Adversarial Generator (LAG). In our generative sampling framework, we only use the input (possibly of very low- resolution) to direct what class of samples the network should produce. As such, the output of our algorithm is not a unique image that relates to the input, but rather a possible {\em set} of related images sampled from the manifold of natural images. Our method learns exclusively in the latent space of the adversary using perceptual loss -- it does not have a pixel loss. 
\end{abstract}

%%%%%%%%% BODY TEXT

\section{Introduction}

We are concerned with the task of generating high resolution (HR) images. In the context of super-resolution from a low resolution (LR) input, this task has been researched extensively and the current best methods are based on deep learning (DL). In the deep learning setting, single image super-resolution is modeled as a regression problem. The neural network weights are optimized to minimize a loss representing the distance from the predicted image to the ground truth. This is {\em not} the aim of this paper. We do not care for generating an image that is close to input (when down-sampled), but rather we  wish to use the input as {\em guidance} toward generating a set of plausible high resolution images. 

To this end, we need a different notion of closeness. Indeed, the concept of distance between images has recently evolved beyond the classical regression framework. Consider two very different applications: In \cite{gatys2015texture}, the authors introduced a new pairwise distance computed in a high level of abstraction space inferred from an inception classifier layer. This distance is associated with a content loss, which, when minimized, results in better modeling of the semantic content but also in visual artifacts. Meanwhile more recently, SRGAN \cite{ledig2017photo} added a new distributional loss term by using a Generative Adversarial Network (GAN) \cite{goodfellow2014generative}. This new loss term is used to capture the distributional properties of image patches which, in practice, results in sharper images.

For image reconstruction purposes (such as denoising and super-resolution), the various distances (pixel, content, distributional) are typically combined together to form an overall loss that includes perceptual measures. Still more losses can be added, for example a texture loss \cite{sajjadi2017enhancenet} or a contextual loss \cite{mechrez2018learning}, leading to various perception-distortion trade-offs \cite{blau2018perception}. The perceptual component of the loss is often manually adjusted against other losses; this requiring extensive fiddling to determine the right balance. This problem has been addressed in \cite{wang2018esrgan} by interpolating between neural networks parameters to manually find the right amount of sharpness; yet it is still not an automated process.

\subsection{Contributions}

In this paper, we propose a novel approach to generating high resolution images, guided by small inputs,  that results in perceptually convincing details. To accomplish this, we break with the previous approaches and seek a single perceptual latent space that encompasses all the desirable properties of the resulting sampled images without manually setting weights.

Our proposed method, called Latent Adversarial Generator (LAG), aims to address the aforementioned fundamental limitations of existing techniques. We present the following contributions:

\begin{itemize}
    \item We model the input images as a set of possibilities rather than a single choice. This in effect models the manifold of (low-resolution) input images. 
    \item We learn a single perceptual latent space in which to describe distances between prediction and ground truth.
    \item We analyze the relationship between conditional GANs and LAG.
\end{itemize}

\section{Latent Adversarial Generator}
Given a low resolution input image $y$, we want to predict the perceptual center $x$ of possible high resolution images.
We propose to achieve this by modeling the possible choices as a random vector $z\in\mathbb{R}^n, z\sim\mathcal{N}(0,1)$.
In this model, a pair $(y, z)$ uniquely maps to a high resolution image $x_z$.
We make the assumption that the high resolution image $x$ is obtained at the center of the normal distribution for $z=0$.

The function we train takes a pair $(y, z)$ and predicts a high resolution image $x_z$.
We adopt the GANs terminology and call this function $G$ a {\em generator}; it has the following signature:
\begin{equation}
    G(x, z; \theta): \mathbb{R}^{y}\times \mathbb{R}^{z} \mapsto \mathbb{R}^{x}
\end{equation}

We design the critic function to judge whether a high resolution image $x$ corresponds to a low resolution image $y$.
We propose to decompose the critic $C$ into two functions: $P$ the projection from an image to a latent space $\mathcal{P}$ and $F$ the mapping from this latent space to $\mathbb{R}$.
We will refer to  $\mathcal{P}$ as the perceptual latent space.
Formally, we define the projection function $P$, parameterized by $\phi$ as:
\begin{equation}
    P(x, y; \phi): \mathbb{R}^{x} \times \mathbb{R}^{y} \mapsto \mathcal{P}
\end{equation}
and the critic $C: \mathbb{R}^{x} \times \mathbb{R}^{y} \mapsto \mathbb{R}$ parameterized by $\phi$ and $\psi$ is the composition of $F$ and $P$:
\begin{equation}
    C(x, y; \psi, \phi) = F(P(x, y; \phi); \psi)
\end{equation}

The functions $G$, $P$ and $F$ are implemented using neural networks. For the remainder of this paper, for simplicity of notation, we omit the parameters $\theta$, $\phi$ and $\psi$ of these functions.

\subsection{Training}
Both the generator $G$ and the critic $C$ are learned adversarially by means of a minimax game.
What particular GAN loss is used is not essential for our mathematical formulation.
However it does matter from a quality of results point view.
We decided to use the Wasserstein GAN objective \cite{arjovsky2017wasserstein} since it led to good visual results: 
\begin{equation}
    \min_G\max_{C\in\mathcal{C}}C(x, y) - C(G(y, z), y)
\end{equation}
\noindent where $\mathcal{C}$ is the space of 1-Lipschitz functions. The constraint $C\in\mathcal{C}$ is implemented using a gradient penalty \cite{gulrajani2017improved} loss for the critic:
\begin{equation}
    \mathcal{L}_{\nabla}(C)=(||\nabla_{\hat{x}}C(\hat{x}, y)||_2 - 1||)^2
\end{equation}

where $\hat{x}$ is a uniformly sampled point between $x$ and $G(y,z)$.
Formally $\hat{x}=\rho x+(1-\rho)G(y,z)$ for a random variable $\rho\in \mathcal{U}(0, 1)$.

While distributional alignment is not available for the training of general purpose GANs, it is available for our task of high resolution image generation as follows. The perceptual representation of a high resolution image $x$ is $P(x)$. For the corresponding input image $y$, the perceptual center is $P(G(y, 0))$. We define the generator loss for aligning the centers as:
\begin{equation}
    \mathcal{L}_{center}(G)=\left(P(x, y)-P(G(y,0), y)\right)^2
\end{equation}

To summarize, the losses for the generator and critic are:
\begin{align}
\mathcal{L}(G)&=-C(G(y, z), y) + \mathcal{L}_{center}(G) \\
\mathcal{L}(C)&=C(G(y, z), y) - C(x, y) + \mathcal{L}_{\nabla}(C)
\end{align}

%\subsection{Limitations}\label{sec:limitations}
The simplifying assumption that $x$ lies in the center of the perceptual space, namely at $z=0$, may be considered a limitation of our method. 

This limitation can be ameliorated to some extent by learning an embedding vector $z_i$ with each training sample $x_i$ with the constraint that $z_i\in\mathbb{R}^n, z\sim\mathcal{N}(0,1)$.
This would make the convergence much slower since each embedding vector is updated only when its corresponding training sample appears in the mini-batch.

Another way to overcome this limitation would be to use an embedding function $E: \mathbb{R}^{x} \mapsto \mathbb{R}^n$ parameterized by $\theta_E$ to obtain $z_i=E(x_i; \theta_E)$, such that $z_i\sim\mathcal{N}(0,1)$.
While not as flexible as the discrete embedding approach, it is faster to learn since its weights can be updated with each mini-batch.

These extensions are left for future research.

\subsection{Relation to conditional GANs}
The approach we propose may be conflated with conditional GANs. Let's clarify the difference. Conditional GANs \cite{mirza2014conditional} are GANs in which the generated sample $x$ is conditioned on the input $y$. Specializing to our case, $x$ is a high resolution image and $y$ is its corresponding low resolution image $y$. The resulting conditional GAN formulation is:
\begin{equation}
    \min_G \max_D \log(D(x|y)) + \log(1-D(G(y)|y))
\end{equation}
Generating a high resolution image conditioned on a low resolution image is a continuous conditional GAN problem.
There is, however, a subtle nuance that distinguishes this approach from ours. In our case, the goal for the generated image is to be both plausible, and close to the ground truth (in the latent space). By comparison, in the  continuous conditional GAN framework, the generated image must be down-scalable to {\em precisely} the same $y$ and be indistinguishable from a real image from the discriminator's judgment {\em without} regard for closeness to ground truth.

As such, our proposed approach can be interpreted as a novel form of a conditional GAN, augmented to consider closeness to ground truth. This is achieved with the loss $\mathcal{L}_{centers}$ that minimizes the distance between $x$ and $G(y)$ in the adversarial latent space.

\section{Implementation Choices}

\subsection{Losses, Conditioning and Architecture}
We've used the Wasserstein GAN loss using gradient penalty. It should be noted that we also obtained good results with a relativistic GAN \cite{jolicoeur2018relativistic,wang2018esrgan} loss paired with spectral normalized convolutions. We did not do a full exploration of all possible GANs losses, as it is beyond the scope of this paper.

We simplified the task of the critic by providing it with the absolute difference with the low resolution input  truth rather than the ground truth itself. In other words, we compute:
\begin{equation}
\begin{cases}
C(x, 0)         &\text{for real samples}\\
C(x_z, |y-\lfloor rH(x_z)\rceil|/r) &\text{otherwise.}
\end{cases}
\end{equation}

where $x_z=G(y, z)$ are generated samples and $H: \mathbb{R}^{x} \mapsto \mathbb{R}^{y}$ is the down-scaling operator and $r$ is the color resolution.
The down-scaling operator produces the low resolution image of a corresponding high resolution image.
We round the output of downsampling operator to its nearest color resolution, in our case $r=2/255$ since we represent images on $[-1, 1]$.
This is done to prevent the network from becoming unstable by converging to large weights to measure infinitesimal differences.
To allow gradient propagation through the rounding operation, we use Hinton's straight through estimator \cite{hinton@2012estimator}.
Assuming a stop gradient operation $\mathtt{sg}$, the straight through estimator for rounding a value $x$ is:
\begin{equation}
    \mathtt{sg}(\lfloor x \rceil - x) + x
\end{equation}

We do not advocate a specific neural network architecture as there are a wide variety of other potential implementation candidates. Indeed, newer and better architectures are constantly appearing in the literature \cite{tong2017image,wang2018fully} and LAG should be adaptable to these other architecture.
In practice, for our experiments, we decided to use a residual network similar to EDSR \cite{lim2017enhanced} for its simplicity. For the critic, we used almost the same architecture but in reverse order.

The architecture is trained by progressively growing the network as introduced in \cite{karras2017progressive,wang2018fully}. This training procedure is not required but seems to yield slightly better visual results.

All the implementation details of the architecture and training, as well as the TensorFlow code can be found here
\ifsubmission
 \emph{URL hidden for submission anonymity}.
\else
 \texttt{https://github.com/google-research/lag}.
\fi

\subsection{Creating Small Inputs}

Ideally, for a particular task like super-resolution, a specific down-scaling operator is used to generate low-resolution images. This operator is determined by the physical process that generates such images (e.g. a camera) or by a choice of prior distribution on the set of such images. 

In our case, however, the choice of a specific operator that generates low-res images is entirely irrelevant since our generator operates in the latent space and should create plausible images from \emph{any} low-resolution input. While the choice of a down-scaling operator is an interesting topic of its own \cite{bulat2018learn}, our main goal is generate images and evaluate the quality of our output's robustness to lack of such information. But to be specific, in our results we simply used the bi-cubic down-scaling function and the average pooling function to generate very low resolution images.

As will be seen in the experiments, regardless of the down-scaling choice, our method achieves roughly the same scores on the metrics under consideration, and produces high quality, realistic and plausible outputs. This shows experimentally that our technique is robust to the process for generating input images.

\section{Experimental Results}

Measuring the quality of a generative model is the subject of ongoing research. We employ two different approaches. The main approach we employ to demonstrate the proposed method empirical, based on a broad set of experiments that illustrate the flexibility of our framework.

The other (presented in the Supplementary Material) is quantitative -- based on comparison of perceptual quality against other approaches to generating high res images from low-resolution inputs (what is sometimes called super-resolution). Since solving super-resolution is {\em not} the stated aim of this paper, this quantitative comparison is presented only in the Supplemental Material. In the GAN literature, several metrics have been developed to measure the quality of the generated sample distributions. Unfortunately, these metrics are primarily focused on capturing the sample diversity and there is no standalone perceptual quality estimation currently available. Other metrics \cite{talebi2018nima,ma2017learning} have been proposed to measure the perceptual quality of a single image.

\section{Generating Sets of Plausible Images}
By design, the LAG method's main strength is the ability to generate not just one, but a family of plausible images given a low-resolution input. Namely, while we can model the set of possible images and predict the one at their center, we can also generate samples by drawing from the distribution of $z\sim\mathcal{N}(0,1)$. We illustrate this capacity with examples in three categories: Faces, Churches, and Bedrooms. We also illustrate examples  where the generator is trained on one class, but the input image is taken from a different class.

\begin{figure*}[t]
\begin{center}
\begin{tabular}{c@{\hskip 4pt}c@{\hskip 2pt}c@{\hskip 2pt}c@{\hskip 2pt}c@{\hskip 2pt}c@{\hskip 2pt}c@{\hskip 2pt}c@{\hskip 2pt}c@{\hskip 2pt}c}
\includegraphics[width=0.11\textwidth]{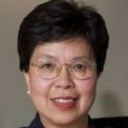} &
\includegraphics[width=0.11\textwidth]{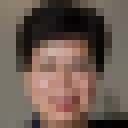} &
\includegraphics[width=0.11\textwidth]{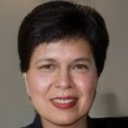} &
\includegraphics[width=0.11\textwidth]{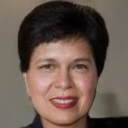} &
\includegraphics[width=0.11\textwidth]{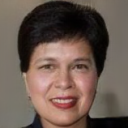} &
\includegraphics[width=0.11\textwidth]{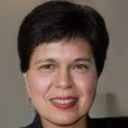} &
\includegraphics[width=0.11\textwidth]{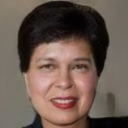} &
\includegraphics[width=0.11\textwidth]{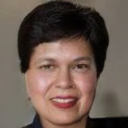} &
\includegraphics[width=0.11\textwidth]{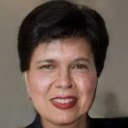} &\\
\includegraphics[width=0.11\textwidth]{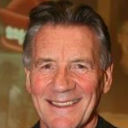} &
\includegraphics[width=0.11\textwidth]{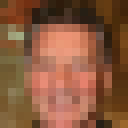} &
\includegraphics[width=0.11\textwidth]{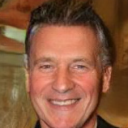} &
\includegraphics[width=0.11\textwidth]{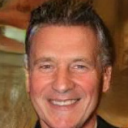} &
\includegraphics[width=0.11\textwidth]{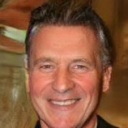} &
\includegraphics[width=0.11\textwidth]{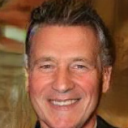} &
\includegraphics[width=0.11\textwidth]{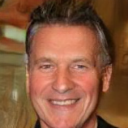} &
\includegraphics[width=0.11\textwidth]{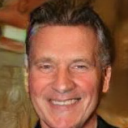} &
\includegraphics[width=0.11\textwidth]{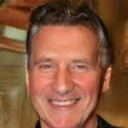} &\\
\includegraphics[width=0.11\textwidth]{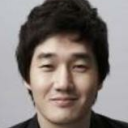} &
\includegraphics[width=0.11\textwidth]{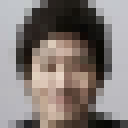} &
\includegraphics[width=0.11\textwidth]{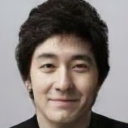} &
\includegraphics[width=0.11\textwidth]{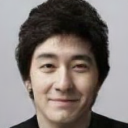} &
\includegraphics[width=0.11\textwidth]{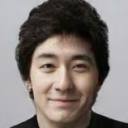} &
\includegraphics[width=0.11\textwidth]{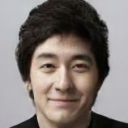} &
\includegraphics[width=0.11\textwidth]{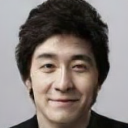} &
\includegraphics[width=0.11\textwidth]{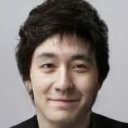} &
\includegraphics[width=0.11\textwidth]{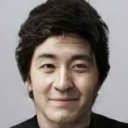} & \\
\includegraphics[width=0.11\textwidth]{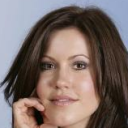} &
\includegraphics[width=0.11\textwidth]{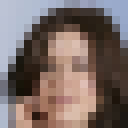} &
\includegraphics[width=0.11\textwidth]{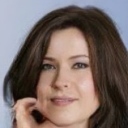} &
\includegraphics[width=0.11\textwidth]{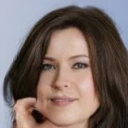} &
\includegraphics[width=0.11\textwidth]{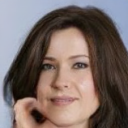} &
\includegraphics[width=0.11\textwidth]{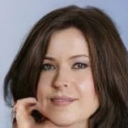} &
\includegraphics[width=0.11\textwidth]{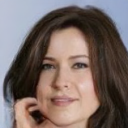} &
\includegraphics[width=0.11\textwidth]{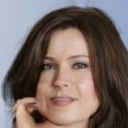} &
\includegraphics[width=0.11\textwidth]{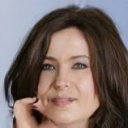} &\\
\text{High res} & \text{Low res} & $z=0$ & \multicolumn{6}{c}{Samples with  $z\sim\mathcal{N}(0,1)$} \\
\end{tabular}
\end{center}
\caption{Samples of generated image from an $8\times$ down-sampled input for various $z$. Note that the various results are similar, relatively consistent, but not identical. Spurious objects such as glasses are not generated.}
\label{fig:zspace8}
\end{figure*}

\begin{figure*}[t]
\begin{center}
\begin{tabular}{c@{\hskip 4pt}c@{\hskip 2pt}c@{\hskip 2pt}c@{\hskip 2pt}c@{\hskip 2pt}c@{\hskip 2pt}c@{\hskip 2pt}c@{\hskip 2pt}c@{\hskip 2pt}c}
\includegraphics[width=0.11\textwidth]{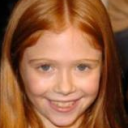} &
\includegraphics[width=0.11\textwidth]{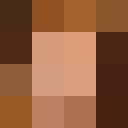} &
\includegraphics[width=0.11\textwidth]{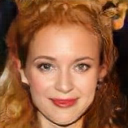} &
\includegraphics[width=0.11\textwidth]{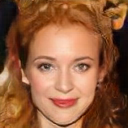} &
\includegraphics[width=0.11\textwidth]{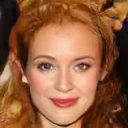} &
\includegraphics[width=0.11\textwidth]{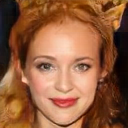} &
\includegraphics[width=0.11\textwidth]{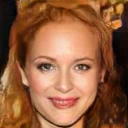} &
\includegraphics[width=0.11\textwidth]{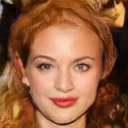} &
\includegraphics[width=0.11\textwidth]{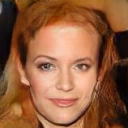} &\\
\includegraphics[width=0.11\textwidth]{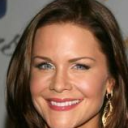} &
\includegraphics[width=0.11\textwidth]{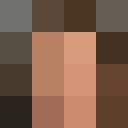} &
\includegraphics[width=0.11\textwidth]{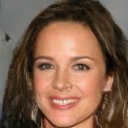} &
\includegraphics[width=0.11\textwidth]{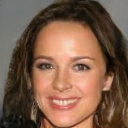} &
\includegraphics[width=0.11\textwidth]{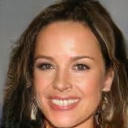} &
\includegraphics[width=0.11\textwidth]{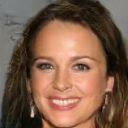} &
\includegraphics[width=0.11\textwidth]{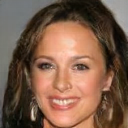} &
\includegraphics[width=0.11\textwidth]{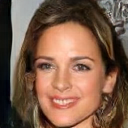} &
\includegraphics[width=0.11\textwidth]{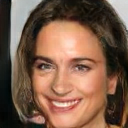} &\\
\includegraphics[width=0.11\textwidth]{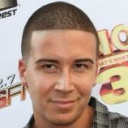} &
\includegraphics[width=0.11\textwidth]{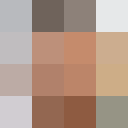} &
\includegraphics[width=0.11\textwidth]{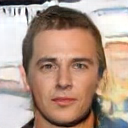} &
\includegraphics[width=0.11\textwidth]{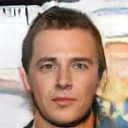} &
\includegraphics[width=0.11\textwidth]{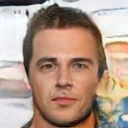} &
\includegraphics[width=0.11\textwidth]{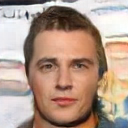} &
\includegraphics[width=0.11\textwidth]{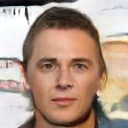} &
\includegraphics[width=0.11\textwidth]{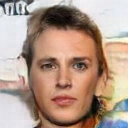} &
\includegraphics[width=0.11\textwidth]{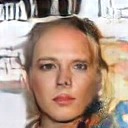} & \\
\includegraphics[width=0.11\textwidth]{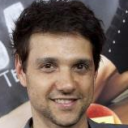} &
\includegraphics[width=0.11\textwidth]{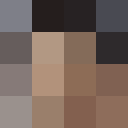} &
\includegraphics[width=0.11\textwidth]{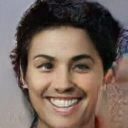} &
\includegraphics[width=0.11\textwidth]{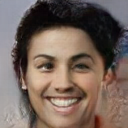} &
\includegraphics[width=0.11\textwidth]{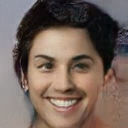} &
\includegraphics[width=0.11\textwidth]{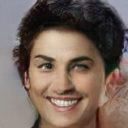} &
\includegraphics[width=0.11\textwidth]{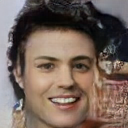} &
\includegraphics[width=0.11\textwidth]{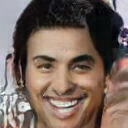} &
\includegraphics[width=0.11\textwidth]{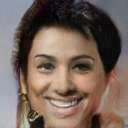} &\\
\includegraphics[width=0.11\textwidth]{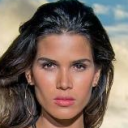} &
\includegraphics[width=0.11\textwidth]{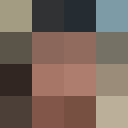} &
\includegraphics[width=0.11\textwidth]{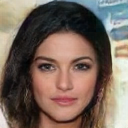} &
\includegraphics[width=0.11\textwidth]{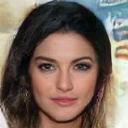} &
\includegraphics[width=0.11\textwidth]{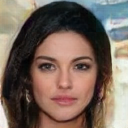} &
\includegraphics[width=0.11\textwidth]{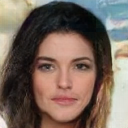} &
\includegraphics[width=0.11\textwidth]{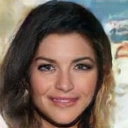} &
\includegraphics[width=0.11\textwidth]{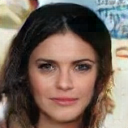} &
\includegraphics[width=0.11\textwidth]{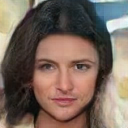} &\\
\text{High res} & \text{Low res} & $z=0$ &\multicolumn{6}{c}{Samples with  $z\sim\mathcal{N}(0,1)$} \\
\end{tabular}
\end{center}
\caption{Samples of generated image from an $32\times$ down-sampled input for various $z$. Note that given the paucity of information in the input, the variance across the resulting generated images is quite large.}
\label{fig:zspace32}
\end{figure*}

We observe in Fig. \ref{fig:zspace8} examples for faces. Specifically, with a modest down-sampling factor, the set of generated images by sampling $z$ is rather similar and consistent. That is, the variance across the samples is modest. However, Fig. \ref{fig:zspace32} shows that by providing the network with a very low resolution image (i.e. high down-sampling factor) the  set of generated images by sampling $z$ is much more varied. This means that the size of the input  in effect controls the diversity of the output. A desirable property that should, in retrospect not be entirely unexpected for a well-designed generator.

\begin{figure*}[t]
\begin{center}
\begin{tabular}{c@{\hskip 4pt}c@{\hskip 2pt}c@{\hskip 2pt}c@{\hskip 2pt}c@{\hskip 2pt}c@{\hskip 2pt}c@{\hskip 2pt}c@{\hskip 2pt}c@{\hskip 2pt}c}
\includegraphics[width=0.11\textwidth]{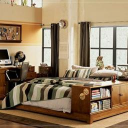} &
\includegraphics[width=0.11\textwidth]{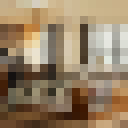} &
\includegraphics[width=0.11\textwidth]{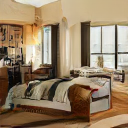} &
\includegraphics[width=0.11\textwidth]{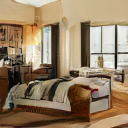} &
\includegraphics[width=0.11\textwidth]{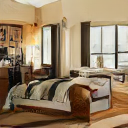} &
\includegraphics[width=0.11\textwidth]{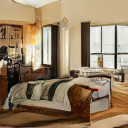} &
\includegraphics[width=0.11\textwidth]{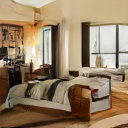} &
\includegraphics[width=0.11\textwidth]{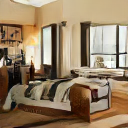} &
\includegraphics[width=0.11\textwidth]{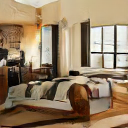} &\\
\includegraphics[width=0.11\textwidth]{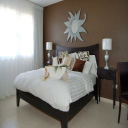} &
\includegraphics[width=0.11\textwidth]{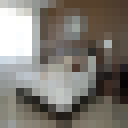} &
\includegraphics[width=0.11\textwidth]{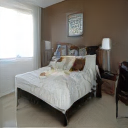} &
\includegraphics[width=0.11\textwidth]{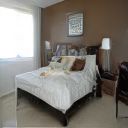} &
\includegraphics[width=0.11\textwidth]{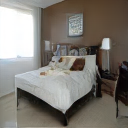} &
\includegraphics[width=0.11\textwidth]{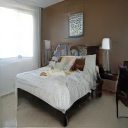} &
\includegraphics[width=0.11\textwidth]{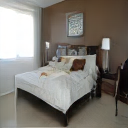} &
\includegraphics[width=0.11\textwidth]{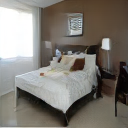} &
\includegraphics[width=0.11\textwidth]{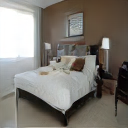} &\\
\includegraphics[width=0.11\textwidth]{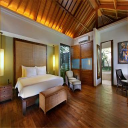} &
\includegraphics[width=0.11\textwidth]{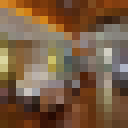} &
\includegraphics[width=0.11\textwidth]{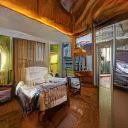} &
\includegraphics[width=0.11\textwidth]{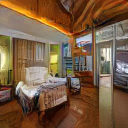} &
\includegraphics[width=0.11\textwidth]{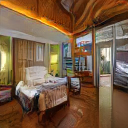} &
\includegraphics[width=0.11\textwidth]{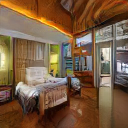} &
\includegraphics[width=0.11\textwidth]{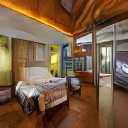} &
\includegraphics[width=0.11\textwidth]{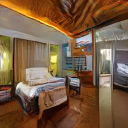} &
\includegraphics[width=0.11\textwidth]{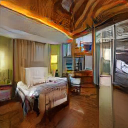} &\\
\includegraphics[width=0.11\textwidth]{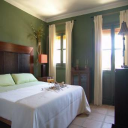} &
\includegraphics[width=0.11\textwidth]{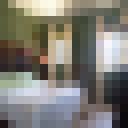} &
\includegraphics[width=0.11\textwidth]{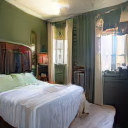} &
\includegraphics[width=0.11\textwidth]{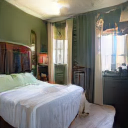} &
\includegraphics[width=0.11\textwidth]{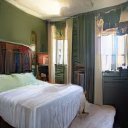} &
\includegraphics[width=0.11\textwidth]{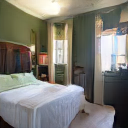} &
\includegraphics[width=0.11\textwidth]{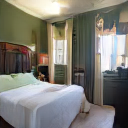} &
\includegraphics[width=0.11\textwidth]{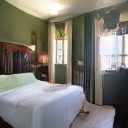} &
\includegraphics[width=0.11\textwidth]{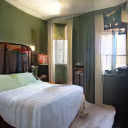} &\\
\text{High res} & \text{Low res} & $z=0$ & \multicolumn{6}{c}{Samples with  $z\sim\mathcal{N}(0,1)$} \\
\end{tabular}
\end{center}
\caption{Samples of generated bedroom images from $8\times$ down-sampled input images for various $z$.}
\label{fig:bedroom16}
\end{figure*}

\begin{figure*}[t]
\begin{center}
\begin{tabular}{c@{\hskip 4pt}c@{\hskip 2pt}c@{\hskip 2pt}c@{\hskip 2pt}c@{\hskip 2pt}c@{\hskip 2pt}c@{\hskip 2pt}c@{\hskip 2pt}c@{\hskip 2pt}c}
\includegraphics[width=0.11\textwidth]{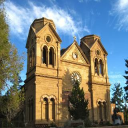} &
\includegraphics[width=0.11\textwidth]{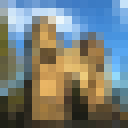} &
\includegraphics[width=0.11\textwidth]{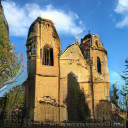} &
\includegraphics[width=0.11\textwidth]{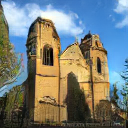} &
\includegraphics[width=0.11\textwidth]{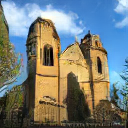} &
\includegraphics[width=0.11\textwidth]{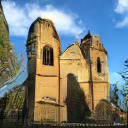} &
\includegraphics[width=0.11\textwidth]{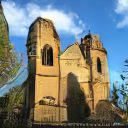} &
\includegraphics[width=0.11\textwidth]{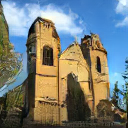} &
\includegraphics[width=0.11\textwidth]{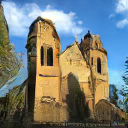} &\\
\includegraphics[width=0.11\textwidth]{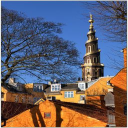} &
\includegraphics[width=0.11\textwidth]{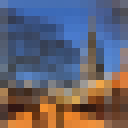} &
\includegraphics[width=0.11\textwidth]{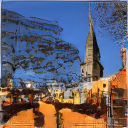} &
\includegraphics[width=0.11\textwidth]{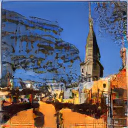} &
\includegraphics[width=0.11\textwidth]{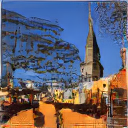} &
\includegraphics[width=0.11\textwidth]{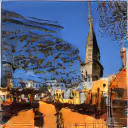} &
\includegraphics[width=0.11\textwidth]{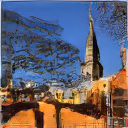} &
\includegraphics[width=0.11\textwidth]{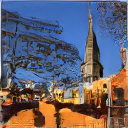} &
\includegraphics[width=0.11\textwidth]{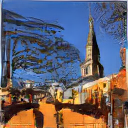} &\\
\includegraphics[width=0.11\textwidth]{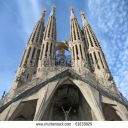} &
\includegraphics[width=0.11\textwidth]{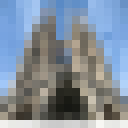} &
\includegraphics[width=0.11\textwidth]{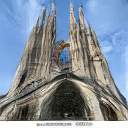} &
\includegraphics[width=0.11\textwidth]{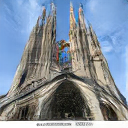} &
\includegraphics[width=0.11\textwidth]{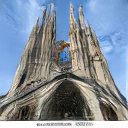} &
\includegraphics[width=0.11\textwidth]{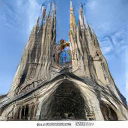} &
\includegraphics[width=0.11\textwidth]{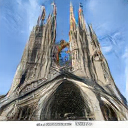} &
\includegraphics[width=0.11\textwidth]{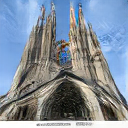} &
\includegraphics[width=0.11\textwidth]{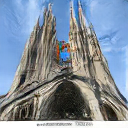} &\\
\includegraphics[width=0.11\textwidth]{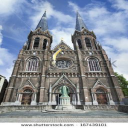} &
\includegraphics[width=0.11\textwidth]{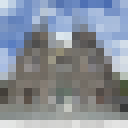} &
\includegraphics[width=0.11\textwidth]{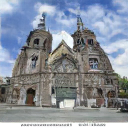} &
\includegraphics[width=0.11\textwidth]{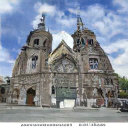} &
\includegraphics[width=0.11\textwidth]{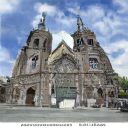} &
\includegraphics[width=0.11\textwidth]{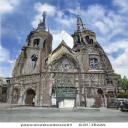} &
\includegraphics[width=0.11\textwidth]{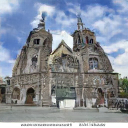} &
\includegraphics[width=0.11\textwidth]{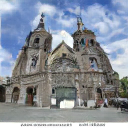} &
\includegraphics[width=0.11\textwidth]{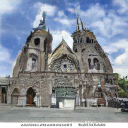} &\\
\text{High res} & \text{Low res} & $z=0$ & \multicolumn{6}{c}{Samples with  $z\sim\mathcal{N}(0,1)$} \\
\end{tabular}
\end{center}
\caption{Samples of generated church images from $8\times$ down-sampled input images for various $z$.}
\label{fig:bedroom16}
\end{figure*}

\subsection{Mirroring}
\begin{figure*}[t]
\begin{center}
\begin{tabular}{c@{\hskip 4pt}c@{\hskip 2pt}c@{\hskip 2pt}c@{\hskip 2pt}c@{\hskip 2pt}c@{\hskip 2pt}c@{\hskip 2pt}c@{\hskip 2pt}c@{\hskip 2pt}c}
\includegraphics[width=0.11\textwidth]{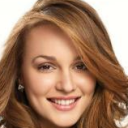} &
\includegraphics[width=0.11\textwidth]{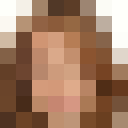} &
\includegraphics[width=0.11\textwidth]{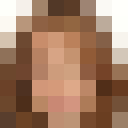} & \includegraphics[width=0.11\textwidth]{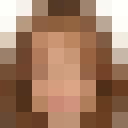} &
\includegraphics[width=0.11\textwidth]{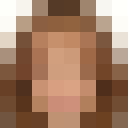} &
\includegraphics[width=0.11\textwidth]{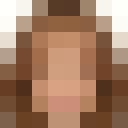} &
\includegraphics[width=0.11\textwidth]{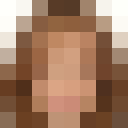} &
\includegraphics[width=0.11\textwidth]{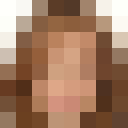} &
\includegraphics[width=0.11\textwidth]{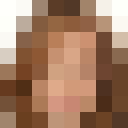} & \\
\includegraphics[width=0.11\textwidth]{samples/mirror/07_00.png} &
\includegraphics[width=0.11\textwidth]{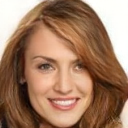} &
\includegraphics[width=0.11\textwidth]{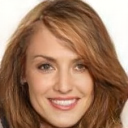} &
\includegraphics[width=0.11\textwidth]{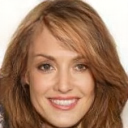} &
\includegraphics[width=0.11\textwidth]{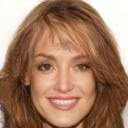} &
\includegraphics[width=0.11\textwidth]{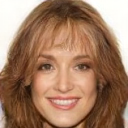} &
\includegraphics[width=0.11\textwidth]{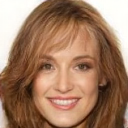} &
\includegraphics[width=0.11\textwidth]{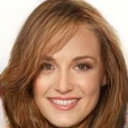} &
\includegraphics[width=0.11\textwidth]{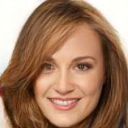} &\\
\includegraphics[width=0.11\textwidth]{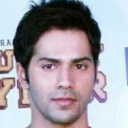} &
\includegraphics[width=0.11\textwidth]{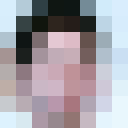} &
\includegraphics[width=0.11\textwidth]{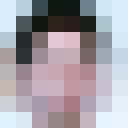} & \includegraphics[width=0.11\textwidth]{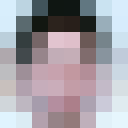} &
\includegraphics[width=0.11\textwidth]{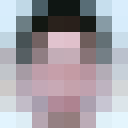} &
\includegraphics[width=0.11\textwidth]{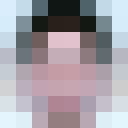} &
\includegraphics[width=0.11\textwidth]{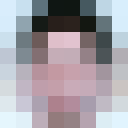} &
\includegraphics[width=0.11\textwidth]{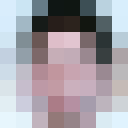} &
\includegraphics[width=0.11\textwidth]{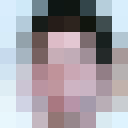} & \\
\includegraphics[width=0.11\textwidth]{samples/mirror/01_00.png} &
\includegraphics[width=0.11\textwidth]{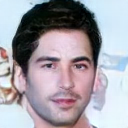} &
\includegraphics[width=0.11\textwidth]{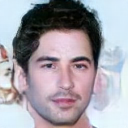} &
\includegraphics[width=0.11\textwidth]{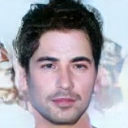} &
\includegraphics[width=0.11\textwidth]{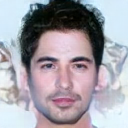} &
\includegraphics[width=0.11\textwidth]{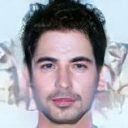} &
\includegraphics[width=0.11\textwidth]{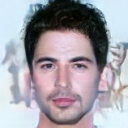} &
\includegraphics[width=0.11\textwidth]{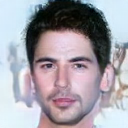} &
\includegraphics[width=0.11\textwidth]{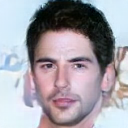} &\\
\end{tabular}
\end{center}
\caption{Samples of generated image from an $16\times$ down-sampled input. The sequence of input images are generated by interpolating between the original high-res and its mirror image. Note that the model generates quite consistent result that illustrate the face going smoothly from a given pose to its mirror image.}
\label{fig:mirror16}
\end{figure*}

In this experiment, we consider how well the network can generate images across a limited, well-defined class. Namely, we consider a given image and it's mirror image. Then we down-sample these images and generated high-resolution versions using our network. The expectation is that the high resolution images should be consistent in the sense that they should illustrate "turning" the given image to its mirror image. Experimental results in \ref{fig:mirror16} bare this out. 

\begin{figure*}[t]
\begin{center}
\begin{tabular}{c@{\hskip 4pt}c@{\hskip 2pt}c@{\hskip 2pt}c@{\hskip 2pt}c@{\hskip 2pt}c@{\hskip 2pt}c@{\hskip 2pt}c@{\hskip 2pt}c@{\hskip 2pt}c}
\includegraphics[width=0.11\textwidth]{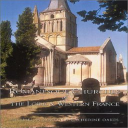} &
\includegraphics[width=0.11\textwidth]{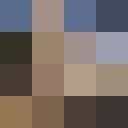} &
\includegraphics[width=0.11\textwidth]{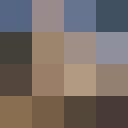} & \includegraphics[width=0.11\textwidth]{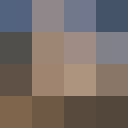} &
\includegraphics[width=0.11\textwidth]{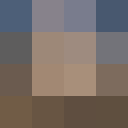} &
\includegraphics[width=0.11\textwidth]{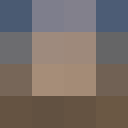} &
\includegraphics[width=0.11\textwidth]{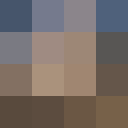} &
\includegraphics[width=0.11\textwidth]{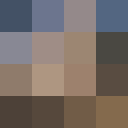} &
\includegraphics[width=0.11\textwidth]{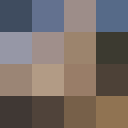} & \\
\includegraphics[width=0.11\textwidth]{samples/mirror/mirrorAB/14_00.png} &
\includegraphics[width=0.11\textwidth]{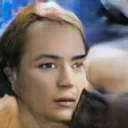} &
\includegraphics[width=0.11\textwidth]{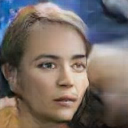} &
\includegraphics[width=0.11\textwidth]{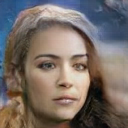} &
\includegraphics[width=0.11\textwidth]{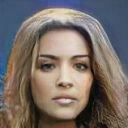} &
\includegraphics[width=0.11\textwidth]{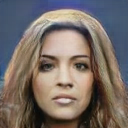} &
\includegraphics[width=0.11\textwidth]{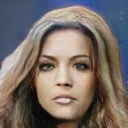} &
\includegraphics[width=0.11\textwidth]{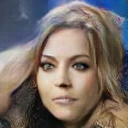} &
\includegraphics[width=0.11\textwidth]{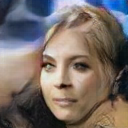} &\\
\includegraphics[width=0.11\textwidth]{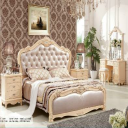} &
\includegraphics[width=0.11\textwidth]{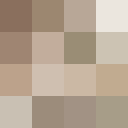} &
\includegraphics[width=0.11\textwidth]{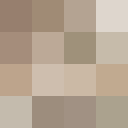} & \includegraphics[width=0.11\textwidth]{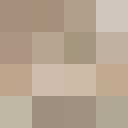} &
\includegraphics[width=0.11\textwidth]{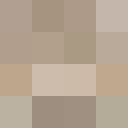} &
\includegraphics[width=0.11\textwidth]{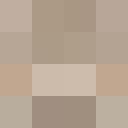} &
\includegraphics[width=0.11\textwidth]{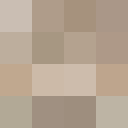} &
\includegraphics[width=0.11\textwidth]{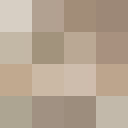} &
\includegraphics[width=0.11\textwidth]{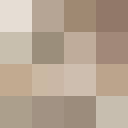} & \\
\includegraphics[width=0.11\textwidth]{samples/mirror/mirrorAB/05_00.png} &
\includegraphics[width=0.11\textwidth]{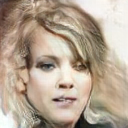} &
\includegraphics[width=0.11\textwidth]{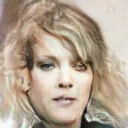} &
\includegraphics[width=0.11\textwidth]{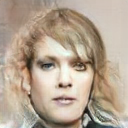} &
\includegraphics[width=0.11\textwidth]{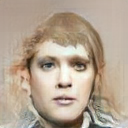} &
\includegraphics[width=0.11\textwidth]{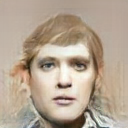} &
\includegraphics[width=0.11\textwidth]{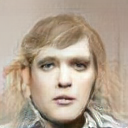} &
\includegraphics[width=0.11\textwidth]{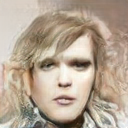} &
\includegraphics[width=0.11\textwidth]{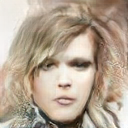} &\\
\end{tabular}
\end{center}
\caption{Samples of generated image from an $16\times$ down-sampled input. The sequence of input images are generated by interpolating between the original high-res and its mirror image. Note that the model generates quite consistent result that illustrate the face going smoothly from a given pose to its mirror image.}
\label{fig:mirrorAB16}
\end{figure*}

Next, we conducted an additional experiment to see how the network behaves when the input image belongs to data-set A and generator is trained on different data-set B. The resuls are shown in \ref{fig:mirrorAB16}

\subsection{Noisy and Random Inputs}

In this experiment, for the sake of completeness, we consider how the network reacts to being given either noisy images, or simply images that consist of only noise. 

\begin{figure*}[t]
\begin{center}
\begin{tabular}{c@{\hskip 4pt}c@{\hskip 2pt}c@{\hskip 2pt}c@{\hskip 2pt}c@{\hskip 2pt}c@{\hskip 2pt}c@{\hskip 2pt}c@{\hskip 2pt}c@{\hskip 2pt}c}
\includegraphics[width=0.11\textwidth]{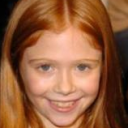} &
\includegraphics[width=0.11\textwidth]{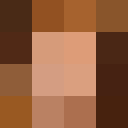} &
\includegraphics[width=0.11\textwidth]{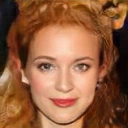} & \includegraphics[width=0.11\textwidth]{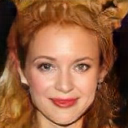} &
\includegraphics[width=0.11\textwidth]{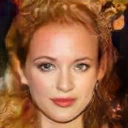} &
\includegraphics[width=0.11\textwidth]{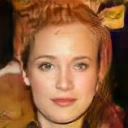} &
\includegraphics[width=0.11\textwidth]{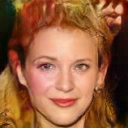} &
\includegraphics[width=0.11\textwidth]{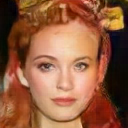} &
\includegraphics[width=0.11\textwidth]{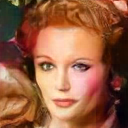} & \\
\includegraphics[width=0.11\textwidth]{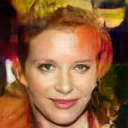} &
\includegraphics[width=0.11\textwidth]{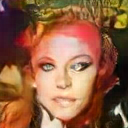} &
\includegraphics[width=0.11\textwidth]{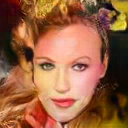} &
\includegraphics[width=0.11\textwidth]{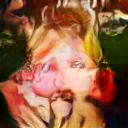} &
\includegraphics[width=0.11\textwidth]{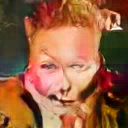} &
\includegraphics[width=0.11\textwidth]{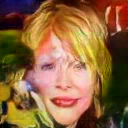} &
\includegraphics[width=0.11\textwidth]{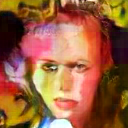} &
\includegraphics[width=0.11\textwidth]{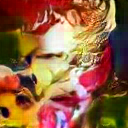} &
\includegraphics[width=0.11\textwidth]{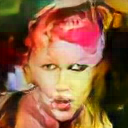} &\\
\\\hline \\
\includegraphics[width=0.11\textwidth]{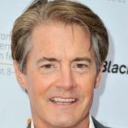} &
\includegraphics[width=0.11\textwidth]{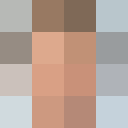} &
\includegraphics[width=0.11\textwidth]{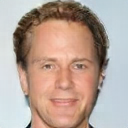} & \includegraphics[width=0.11\textwidth]{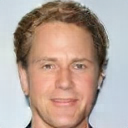} &
\includegraphics[width=0.11\textwidth]{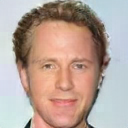} &
\includegraphics[width=0.11\textwidth]{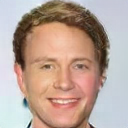} &
\includegraphics[width=0.11\textwidth]{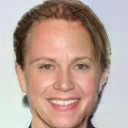} &
\includegraphics[width=0.11\textwidth]{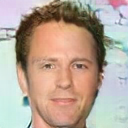} &
\includegraphics[width=0.11\textwidth]{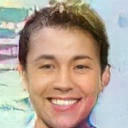} & \\
\includegraphics[width=0.11\textwidth]{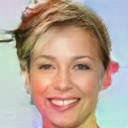} &
\includegraphics[width=0.11\textwidth]{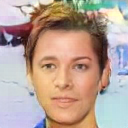} &
\includegraphics[width=0.11\textwidth]{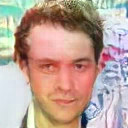} &
\includegraphics[width=0.11\textwidth]{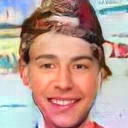} &
\includegraphics[width=0.11\textwidth]{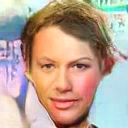} &
\includegraphics[width=0.11\textwidth]{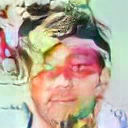} &
\includegraphics[width=0.11\textwidth]{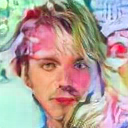} &
\includegraphics[width=0.11\textwidth]{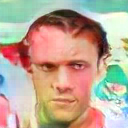} &
\includegraphics[width=0.11\textwidth]{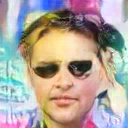} &\\
\end{tabular}
\end{center}
\caption{Two examples of generated image from a $32\times$ down-sampled input with progressively more noise added to the input images. Note that in all cases, the latent parameter $z=0$ was fixed. Top Left: high-res, followed by low-res, and generated images from noisy low-res}
\label{fig:noisy32}
\end{figure*}

It is of interest to see how the  generated set varies for a fixed $z =0$, but where progressively more uniform pixel noise is added to the input image. The results in \ref{fig:noisy32} illustrate a rather varied set of generated outputs.

\section{Conclusion}
We introduced LAG, a new method for generating high resolution images using latent space. It is simple to train and appears robust: we do not need to tune hyper-parameters to avoid mode collapse.

LAG obtains (automatically) through an emergent behavior of the latent space, a good balance between pixel and perceptual accuracy. Our approach results in flexibility in both variations of generated images, and higher quality results when a specific image is to be generated against a reference. This is achieved by training in a perceptual latent space which does not require additional losses to measure pixel or content.

Importantly, LAG allows to explore a space of high resolution images given an input, allowing a flexible framework that seems to allow higher resolution generated images for simpler manifolds. In addition to the latent variable $z$, we identified two additional mechanism for controlling the variety of images the network can generate. First, the size of the input image directly affects the observed variations across images generated by the network. Very small inputs generate a much broader  variety of images, as might be expected. Second, even for a fixed $z$, the addition of modest amounts of noise to the input images is also able to generate a variety of outputs that appear plausible.  

Several avenues for future research seem open. First, alternative architectures can of course be explored. Second, and more fundamentally, our assumption that the reference image lies at the center of a Gaussian distribution is probably a terrible simplification. We expect that even more compelling results are possible.

Last, taking advantage of semantic information that describes the content of an image seems a natural path for producing better high resolution images that may be modeled as coming from simpler manifolds.

\newpage

{\small
\bibliographystyle{ieee}
\bibliography{egbib}

\begin{thebibliography}{10}\itemsep=-1pt

\bibitem{arjovsky2017wasserstein}
M.~Arjovsky, S.~Chintala, and L.~Bottou.
\newblock Wasserstein gan.
\newblock {\em arXiv preprint arXiv:1701.07875}, 2017.

\bibitem{blau2018perception}
Y.~Blau and T.~Michaeli.
\newblock The perception-distortion tradeoff.
\newblock In {\em Proc. 2018 IEEE/CVF Conference on Computer Vision and Pattern
  Recognition, Salt Lake City, Utah, USA}, pages 6228--6237, 2018.

\bibitem{bulat2018learn}
A.~Bulat, J.~Yang, and G.~Tzimiropoulos.
\newblock To learn image super-resolution, use a gan to learn how to do image
  degradation first.
\newblock {\em arXiv preprint arXiv:1807.11458}, 2018.

\bibitem{gatys2015texture}
L.~Gatys, A.~S. Ecker, and M.~Bethge.
\newblock Texture synthesis using convolutional neural networks.
\newblock In {\em Advances in Neural Information Processing Systems}, pages
  262--270, 2015.

\bibitem{goodfellow2014generative}
I.~Goodfellow, J.~Pouget-Abadie, M.~Mirza, B.~Xu, D.~Warde-Farley, S.~Ozair,
  A.~Courville, and Y.~Bengio.
\newblock Generative adversarial nets.
\newblock In {\em Advances in neural information processing systems}, pages
  2672--2680, 2014.

\bibitem{gulrajani2017improved}
I.~Gulrajani, F.~Ahmed, M.~Arjovsky, V.~Dumoulin, and A.~C. Courville.
\newblock Improved training of wasserstein gans.
\newblock In {\em Advances in Neural Information Processing Systems}, pages
  5767--5777, 2017.

\bibitem{hinton@2012estimator}
G.~Hinton.
\newblock Neural networks for machine learning.
\newblock Coursera, video lectures., 2012.

\bibitem{jolicoeur2018relativistic}
A.~Jolicoeur-Martineau.
\newblock The relativistic discriminator: a key element missing from standard
  gan.
\newblock {\em arXiv preprint arXiv:1807.00734}, 2018.

\bibitem{karras2017progressive}
T.~Karras, T.~Aila, S.~Laine, and J.~Lehtinen.
\newblock Progressive growing of gans for improved quality, stability, and
  variation.
\newblock {\em arXiv preprint arXiv:1710.10196}, 2017.

\bibitem{ledig2017photo}
C.~Ledig, L.~Theis, F.~Husz{\'a}r, J.~Caballero, A.~Cunningham, A.~Acosta,
  A.~P. Aitken, A.~Tejani, J.~Totz, Z.~Wang, et~al.
\newblock Photo-realistic single image super-resolution using a generative
  adversarial network.
\newblock In {\em CVPR}, volume~2, page~4, 2017.

\bibitem{lim2017enhanced}
B.~Lim, S.~Son, H.~Kim, S.~Nah, and K.~M. Lee.
\newblock Enhanced deep residual networks for single image super-resolution.
\newblock In {\em The IEEE conference on computer vision and pattern
  recognition (CVPR) workshops}, volume~1, page~4, 2017.

\bibitem{ma2017learning}
C.~Ma, C.-Y. Yang, X.~Yang, and M.-H. Yang.
\newblock Learning a no-reference quality metric for single-image
  super-resolution.
\newblock {\em Computer Vision and Image Understanding}, 158:1--16, 2017.

\bibitem{mechrez2018learning}
R.~Mechrez, I.~Talmi, F.~Shama, and L.~Zelnik-Manor.
\newblock Learning to maintain natural image statistics.
\newblock {\em arXiv preprint arXiv:1803.04626}, 2018.

\bibitem{mirza2014conditional}
M.~Mirza and S.~Osindero.
\newblock Conditional generative adversarial nets.
\newblock {\em arXiv preprint arXiv:1411.1784}, 2014.

\bibitem{sajjadi2017enhancenet}
M.~S. Sajjadi, B.~Sch{\"o}lkopf, and M.~Hirsch.
\newblock Enhancenet: Single image super-resolution through automated texture
  synthesis.
\newblock In {\em Computer Vision (ICCV), 2017 IEEE International Conference
  on}, pages 4501--4510. IEEE, 2017.

\bibitem{talebi2018nima}
H.~Talebi and P.~Milanfar.
\newblock Nima: Neural image assessment.
\newblock {\em IEEE Transactions on Image Processing}, 27(8):3998--4011, 2018.

\bibitem{tong2017image}
T.~Tong, G.~Li, X.~Liu, and Q.~Gao.
\newblock Image super-resolution using dense skip connections.
\newblock In {\em Computer Vision (ICCV), 2017 IEEE International Conference
  on}, pages 4809--4817. IEEE, 2017.

\bibitem{wang2018esrgan}
X.~Wang, K.~Yu, S.~Wu, J.~Gu, Y.~Liu, C.~Dong, C.~C. Loy, Y.~Qiao, and X.~Tang.
\newblock Esrgan: Enhanced super-resolution generative adversarial networks.
\newblock {\em arXiv preprint arXiv:1809.00219}, 2018.

\bibitem{wang2018fully}
Y.~Wang, F.~Perazzi, B.~McWilliams, A.~Sorkine-Hornung, O.~Sorkine-Hornung, and
  C.~Schroers.
\newblock A fully progressive approach to single-image super-resolution.
\newblock {\em arXiv preprint arXiv:1804.02900}, 2018.

\end{thebibliography}
}

\end{document}